\pdfoutput=1

\documentclass[11pt]{article}

\usepackage[final]{acl}

\usepackage{times}
\usepackage{latexsym}

\usepackage[T1]{fontenc}

\usepackage[utf8]{inputenc}

\usepackage{microtype}

\usepackage{inconsolata}

\usepackage{graphicx}

\usepackage{subfig}
\usepackage{amsmath}
\usepackage{amsfonts}
\usepackage{float}
\usepackage{float}
\usepackage{footnote}
\usepackage{enumitem}
\usepackage{bm}
\usepackage{arydshln}
\usepackage{listings}
\usepackage{multicol}
\usepackage{xcolor}
\usepackage{colortbl}
\usepackage{makecell}
\usepackage{mathtools}
\usepackage{imakeidx}
\makeindex
\usepackage{lipsum}
\usepackage{natbib}
\usepackage[toc]{multitoc}
\usepackage[edges]{forest}
\usepackage[normalem]{ulem}
\usepackage{hyperref}

\usepackage{csquotes}
\usepackage{CJKutf8}
\usepackage{awesomebox}
\usepackage{tcolorbox}
\usepackage{soul}

\usepackage{url}
\usepackage{booktabs}

\usepackage{tabularx}
\usepackage{geometry}
\usepackage{multirow, makecell}

\usepackage{caption}
\usepackage{subcaption}

\usepackage{xspace}
\newcommand{\ie}{\textit{i.e.}\xspace}
\newcommand{\eg}{\textit{e.g.}\xspace}

\newcommand{\chemprotchem}{ChemProt-Chem}
\newcommand{\chemprotgene}{ChemProt-Gene}
\newcommand{\bctwogm}{BC2GM}
\newcommand{\bcfivechem}{BC5-Chem}
\newcommand{\bcfivedisease}{BC5-Disease}

\newcommand{\knn}{$k$NN}
\newcommand{\llama}{Llama2}
\newcommand{\gptthree}{GPT-3.5-Turbo}
\newcommand{\picl}{PICLe}
\newcommand{\kmeans}{$k$-means}
\newcommand{\spkmeans}{Sp-$k$-means}

\newcommand\pickle{\raisebox{-8pt}{\includegraphics[trim={15em 1em 15em 1em},clip,width=1.8em]{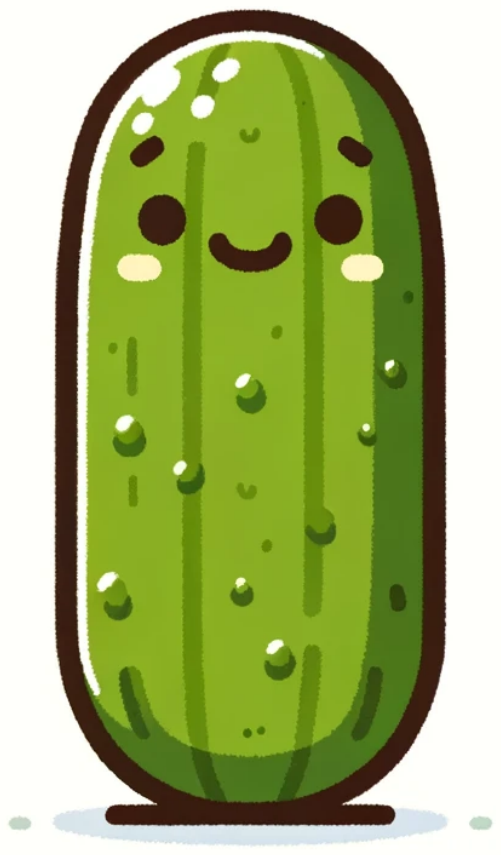}}}

\definecolor{mydarkblue}{rgb}{0,0.08,0.45}
\definecolor{openaigreen}{RGB}{85, 180, 129}

\newtcolorbox{promptbox}{
    colback=gray!10,        
    colframe=gray!80,       
    fonttitle=\bfseries,
    boxrule=0.5pt,
    width=\linewidth,   
    boxsep=5pt,
    arc=10pt,               
    before skip=10pt,       
    after skip=10pt,        
}

\newcommand{\prompt}[1]{
    \begin{promptbox}
    #1
    \end{promptbox}
}

\definecolor{darkblue}{rgb}{0, 0, 0.5}
\hypersetup{colorlinks=true, citecolor=darkblue, linkcolor=darkblue, urlcolor=darkblue}

\title{\pickle~PICLe: Pseudo-Annotations for In-Context Learning \\ in Low-Resource Named Entity Detection}

\author{Sepideh Mamooler, Syrielle Montariol, Alexander Mathis\thanks{Equal Supervision}, Antoine Bosselut$^*$
\\ EPFL \\
\texttt{\{firstname.lastname\}@epfl.ch}
}

\begin{document}
\maketitle
\begin{abstract}
In-context learning~(ICL) enables Large Language Models~(LLMs) to perform tasks using few demonstrations, facilitating task adaptation when labeled examples are hard to obtain.
However, ICL is sensitive to the choice of demonstrations, and it remains unclear which demonstration attributes enable in-context generalization.
In this work, we conduct a perturbation study of in-context demonstrations for low-resource Named Entity Detection~(NED).
Our surprising finding is that in-context demonstrations with partially correct annotated entity mentions can be as effective for task transfer as fully correct demonstrations. 
 
Based off our findings, we propose \textbf{P}seudo-annotated \textbf{I}n-\textbf{C}ontext \textbf{Le}arning~(\picl), a framework for in-context learning with noisy, pseudo-annotated demonstrations. \picl{} leverages LLMs to annotate many demonstrations in a zero-shot first pass. We then cluster these synthetic demonstrations, sample specific sets of in-context demonstrations from each cluster, and predict entity mentions using each set independently. Finally, we use self-verification to select the final set of entity mentions. We evaluate \picl{} on five biomedical NED datasets and show that, with zero human annotation, \picl{} outperforms ICL in low-resource settings where limited gold examples can be used as in-context demonstrations.\footnote{Our code can be found at \url{https://github.com/sMamooler/PICLe}.}

\end{abstract}

\section{Introduction}

With in-context learning~(ICL), Large Language Models~(LLMs) can be adapted to perform many tasks using few demonstrations~\citep{Brown2020LanguageMA,dong2022survey,srivastava2023beyond,ye2023amadeusgpt}. This emergent property of LLMs is particularly beneficial in tasks where limited supervision data is available for fine-tuning models, such as in specialized domains where only expensive expert annotations can be relied upon to produce quality data (\eg, biomedical, clinical, and legal domains, among many others), and in situations where in-house proprietary datasets must be compiled with few available experts to perform the annotation.

Despite its promise in these settings, ICL is highly sensitive to the choice of the demonstrations \citep{Wang2023LargeLM,li-qiu-2023-mot,Liu2021WhatMG}, and it remains unclear which characteristics of demonstrations are critical for successful task adaptation. Consequently, prior work has explored which demonstration characteristics lead to successful task adaptation in ICL~\citep{min-etal-2022-rethinking,yoo-etal-2022-ground,Wei2023LargerLM}, but these studies have focused on scalar-output tasks with a limited, predefined label space, such as classification. As a result, demonstration characteristics that optimize performance remain unclear for tasks that require structured, open-ended prediction such as Named Entity Detection~(NED).

In NED, the goal is to identify all mentions of a specific type of entity within a given query, producing a structured output, with a label space effectively bounded only by the number of domain entities. In this work, we focus on NED given its high number of use cases~\citep{navarro2023clinical,skylaki2020named,ushio-etal-2022-named}, particularly in specialized domains where effective annotation is challenging, as (1) it requires considerable domain expertise, and (2) entities can change over time, introducing distribution shifts in supervised datasets.

We conduct a thorough analysis of demonstration properties that impact in-context adaptation in NED. First, we analyze the importance of the context-label correspondence of in-context demonstrations by introducing noise through various perturbations that preserve different aspects of this mapping. Second, we investigate the effects of \textit{partial correctness} in demonstrations. In NED, partial correctness refers to annotations that differ from the complete list of entity mentions in the input query but still share some overlap with it. To explore this, we apply various perturbation schemes to produce demonstrations with different levels of correctness. We find that while a semantic mapping between the demonstration context and label is essential, even weak semantic mappings can provide sufficient signals for in-context task transfer. Moreover, our analysis reveals that in-context learning is surprisingly resilient to partially correct annotations, provided that the demonstration labels include a large number of entities.

Following this analysis, we introduce \textbf{P}seudo-annotated \textbf{I}n-\textbf{C}ontext \textbf{Le}arning~(\picl), a framework for in-context NED with pseudo-annotated demonstrations that requires no human labeling effort.
First, we exploit a pool of unlabeled samples to obtain pseudo-annotations through zero-shot prediction from LLMs, followed by a self-verification step in which the model is prompted to verify the type of individual entities.
Then, the (noisy) pseudo-annotated samples are clustered, and demonstration sets are sampled from each cluster individually.
These cluster-specific demonstrations are used to predict the entities mentioned in the test query. Predictions from all clusters are consolidated to obtain the final set of entity mentions. 

In our evaluations with multiple LLMs across $5$ biomedical entity detection datasets~\citep{Taboureau2010ChemProtAD,Li2016BioCreativeVC,Smith2008OverviewOB}, we show that \picl{} is as effective as, and on average outperforms, standard ICL that uses gold-labeled demonstrations.

\paragraph{Contributions.} In summary,
\begin{itemize}[nosep]
    \item [1.] we conduct a perturbation study to identify demonstration attributes that enable ICL in low-resource NED. We find that above a surprisingly low correctness threshold, partially correct annotations are as effective for ICL as demonstrations with fully correct gold annotations.
    
    \item [2.] we propose \picl, a novel framework for in-context learning that uses pseudo-annotated demonstrations as in-context examples. 
    We show that without human-annotation effort, \picl{} competes and even outperforms ICL with gold-labeled demonstrations in resource-scarce settings.
\end{itemize}
\section{Related Work}

\textbf{What matters in in-context learning?} 
In-context learning is remarkably effective for performing various NLP tasks with only a few task demonstrations appended to the prompt \citep{Brown2020LanguageMA}. 
However, despite a large body of work on designing novel in-context learning methods~(\eg, \citealp{gao-etal-2021-making,sorensen-etal-2022-information,mishra-etal-2022-reframing}), it is not yet fully understood what makes in-context learning effective, with multiple works demonstrating surprising variables, such as the impact of the demonstration order \citep{lu-etal-2022-fantastically}, the term frequencies of test examples in pretraining data \citep{razeghi-etal-2022-impact}, and basic output calibration \citep{Zhao2021CalibrateBU,fei-etal-2023-mitigating,jiang-etal-2023-generative}. Consequently, recent works explore \textit{how} demonstration components might be separately responsible for in-context transfer. \citet{min-etal-2022-rethinking} show that in-context demonstrations serve to show the label space of demonstrations, the distribution of their input text, and their overall format. However, ~\citet{yoo-etal-2022-ground} perform quantifiable analysis on the impact of ground-truth label demonstrations on a larger set of tasks and datasets and find that ground-truth labels have substantial impacts on ICL performance. \citet{Wei2023LargerLM} continue this line of work and show that the degree to which the label mapping influences task transfer depends on the scale of the model, and that smaller models are more capable of ignoring misaligned label mappings. \citet{wang-etal-2023-towards} show similar results for CoT reasoning, finding that CoT is also possible without valid demonstrations, and that demonstrations that are relevant to the query and have the correct order of reasoning steps are more important for effective transfer. 

However, these works focus on classification tasks, which lack the concept of partial correctness; a label is either fully correct or entirely incorrect. In token-level tasks like NED, however, the list of annotated entities can be partially correct. We show that partially correct demonstrations can perform as effectively as fully correct ones—a result not addressed by these prior works. Furthermore, contrary to \citet{min-etal-2022-rethinking}’s findings for classification tasks, we show that ICL demonstrations with fully incorrect labels are not effective in NED.

\begin{table*}[t]
    \centering
    \resizebox{\linewidth}{!}{ 
    \begin{tabular}{lllcccc}
    \toprule
        \multirow{2}{*}{\textbf{Source}} & \multirow{2}{*}{\textbf{Name}} & \multirow{2}{*}{\textbf{Entity type(s)}} & \multirow{2}{*}{\textbf{\#Train}} & \multirow{2}{*}{\textbf{\#Test}} & \multirowcell{2}[0pt][c]{\textbf{Avg \# words} \\ \textbf{per entity}} & \multirowcell{2}[0pt][c]{\textbf{Ratio null} \\ \textbf{samples (\%)}}\\
          &  &  &  &  &  &  \\
        \midrule
       \multirowcell{2}[0pt][l]{ChemProt\\\citep{Taboureau2010ChemProtAD}} & \chemprotchem & chemical &\multirow{2}{*}{10,732} & \multirow{2}{*}{8,431} & 1.39 & 41.3 \\
         & \chemprotgene & gene/protein & &  & 1.62 & 45.0\\
        \midrule
        \multirowcell{2}[0pt][l]{BC5CDR\\\citep{Li2016BioCreativeVC}} & \bcfivechem & chemical & \multirow{2}{*}{4,560}  & \multirow{2}{*}{4,797} & 1.36 & 35.3\\
        & \bcfivedisease & disease/illness & & & 1.70 & 41.7\\
        \midrule
        \multicolumn{2}{l}{BC2GM~\citep{Smith2008OverviewOB}}  & gene/protein & 12,575 & 5,039 & 2.45 & 48.9 \\
        \bottomrule
    \end{tabular}}
    \caption{\textbf{Datasets' description and statistic}: number of samples~(sentences) in train and test splits, average number of words per entity and null samples~(samples with no labeled entities) ratio in train split. We use the versions available in the HuggingFace library.\protect\footnotemark}
    \label{tab:dataset_info}
\end{table*}

\paragraph{Pseudo-annotation.}
Pseudo-annotation is a popular semi-supervised learning method in many domains~\citep{yang2022survey,ye2024superanimal}. It has recently been used for various NLP tasks to generate demonstrations for ICL~\citep{wan-etal-2023-better,wan-etal-2023-universal} and fine-tuning LLMs~\citep{huang-etal-2023-large,honovich-etal-2023-unnatural,wang-etal-2023-self-instruct}. Demonstrations' pseudo-annotations are either random (\textit{e.g.} Z-ICL, ~\citealp{lyu-etal-2023-z}, for classification tasks) or partially correct~\citep{wan-etal-2023-better,chen-etal-2023-self}. In particular, COSP~\citep{wan-etal-2023-better} selects and builds a demonstration pool from an LLM's zero-shot outputs via multiple rounds of prediction with high temperature and exceeds few-shot baselines for a range of reasoning tasks. Most similar to our work is Self-ICL \citep{chen-etal-2023-self}, which uses zero-shot models to generate in-context demonstrations for text classification. In our work, we construct a pipeline for leveraging zero-shot predicted labels for real test examples in named entity detection, but ground our pseudo-annotation method in analysis of how demonstration noise influences downstream in-context learning performance. 

\paragraph{Information extraction with ICL.}
Although LLMs have achieved SOTA performance in many tasks, their performance in information extraction is still significantly below supervised baselines~\citep{ma-etal-2023-large}. Recent works have designed dedicated prompting techniques to improve in-context NER for LLMs~\citep{lee-etal-2022-good,shen-etal-2023-promptner,chen-etal-2023-learning}.
Prompt-NER~\citep{shen-etal-2023-promptner} provides entity definitions to the model, and prompts it for a list of potential entities with an explanation justifying the compatibility of each entity with the provided definition. Their approach outperforms vanilla prompting, but still requires human effort to annotate gold demonstrations that may not be available in many application settings.
In our work, we adopt a similar task formulation as Prompt-NER, but do not require labeled demonstrations or explanations, as we use pseudo-annotation to produce in-context learning examples.
\begin{table*}[t]
    \centering
    \footnotesize
    \begin{tabular}{lp{0.6\textwidth}}
    \toprule
         \textbf{\texttt{Text}} &  \textsf{This pretreatment had no effect on the inhibition of GABA-T or the elevation of brain GABA levels produced by VIG .} \\
         \textbf{\texttt{Gold Labels}} & [\textsf{GABA, GABA, VIG}] \\ 
         \midrule
         \textbf{\texttt{Random ID Labels}} & [\textsf{quinoxalines, W13, N-acetylcysteine}]\\
         \textbf{\texttt{Swapped ID Labels}} & [\textsf{AMG}] \\
          \textbf{\texttt{Random OOD Labels (from nltk})} & [\textsf{unmeliorated, suddy, vista}]\\
          \textbf{\texttt{Random OOD Labels from Text}} & [\textsf{brain, pretreament, elevation}]\\
         \midrule
         \textbf{\texttt{Corrupted OOD Text}} & \textsf{This pretreatment had no effect on the inhibition of \textbf{unmeliorated} or the elevation of brain \textbf{suddy} levels produced by \textbf{vista}.}\\
         \textbf{\texttt{Corrupted and Shuffled OOD Text}} & \textsf{of had by produced on elevation no . levels or effect the \textbf{vista} of the This \textbf{unmeliorated} pretreatment brain inhibition \textbf{suddy}} \\
    \bottomrule
    \end{tabular}
    \caption{\textbf{Examples of different text and labels corruption schemes}. Source: \chemprotchem.}
    \label{tab:incorrect_demo_example}
\end{table*}

\section{Experimental setup}\label{sec:exp-setup}

The task of Named Entity Detection~(NED) requires detecting all mentions of entities in a text. We formulate the task such that the language model is given a passage of text as part of a prompt and must predict the list of entities that are mentioned in the passage. Optionally, in few-shot settings (\ie, in-context learning), the prompt also contains several demonstrations, each including an example passage and a corresponding list of mentioned entities in the passage.

\paragraph{Datasets.} 
We consider five biomedical NED datasets with rich and comprehensive collections of diverse specialized entity types~(Table~\ref{tab:dataset_info}).
\textbf{ChemProt} contains annotations for extracting chemical compounds~(drugs) and gene and protein-related objects~\citep{Taboureau2010ChemProtAD}. Originally, each sample of this dataset is a paragraph, but we split these paragraphs into sentences. We construct two datasets from ChemProt: \textit{\chemprotchem} and \textit{\chemprotgene}, for detecting chemicals and genes, respectively. \textbf{BC5CDR} contains biomedical abstracts annotated for chemical and disease extraction~\citep{Li2016BioCreativeVC}. Similar to ChemProt, we conduct our experiments on two sub-portions, \textit{\bcfivechem} and \textit{\bcfivedisease}. Finally, \textbf{BC2GM} contains biomedical abstracts annotated for the extraction of genes, proteins, and related entities~\citep{Smith2008OverviewOB}.

\footnotetext{\url{https://huggingface.co/bigbio}}

\paragraph{Models.}  
We use three LLMs in our experiments: the proprietary \gptthree{} model, and the open-source Mistral-7b-instruct~\citep{Jiang2023Mistral7} and Llama-2-7b-Chat ~\citep{touvron2023llama} models. In the remainder of the paper, we refer to them as Mistral and \llama{}, respectively.

\paragraph{Metrics.}  
Using each dataset's original IOB$2$~(Inside-Outside-Beginning) annotation scheme, we compute the micro-averaged precision, recall, and F1 score to measure entity mention detection performance.\footnote{We use \texttt{sequeval} (\url{https://github.com/chakki-works/seqeval/}) a widely-used Python library for sequence labeling evaluation.} We consider an entity as correctly detected only if the model extracts the exact span: if some tokens are added or missing compared to the gold span, it is marked incorrect.

\section{Do we need gold demonstrations?}
\label{sec:What_Matters_in_In-context_NER}

In this section, we analyze which components of ICL demonstrations are critical for task transfer by studying the effect of fully incorrect~(Section~\ref{sec:fully_incorrect_demos}) and partially incorrect~(Section~\ref{sec:partial_correctness}) demonstrations in the in-context prompt. In all analyses, we use \knn{} demonstration retrieval~\citep{liu-etal-2022-makes}.\footnote{A comparison of different retrieval methods is provided in Appendix~\ref{app:results}, Figure~\ref{fig:ablation_gold}.}

\subsection{Input-output correspondence of in-context demonstrations}\label{sec:fully_incorrect_demos}

Prior research shows that correct demonstrations are not imperative for priming models in classification tasks \citep{lyu-etal-2023-z}, and that incorrect demonstrations are sufficient to show desired in-context transfer behavior, including domain relevance and annotation format.

In our analysis, we investigate essential demonstration attributes for successful in-context task transfer in NED by designing various corruption schemes, each targeting specific demonstration aspects (see Table~\ref{tab:incorrect_demo_example} for examples). We compare performance under these corruptions to zero-shot prediction (\textbf{\texttt{No Demo}}) and standard ICL (\textbf{\texttt{Gold Label}}). We compare all settings using the same test samples and instructions prepended to the prompt.

\noindent \textbf{\texttt{Random ID Label}:} We replace ground-truth entity labels with random in-distribution entities. For each input sentence, every entity in the ground-truth annotation is replaced by an in-distribution~(ID) entity randomly sampled from all labels in training samples of the dataset. 

\noindent \textbf{\texttt{Swapped ID Labels}:} We swap entity labels in the ground-truth demonstrations with the entity labels of a randomly chosen sample in the training split. Contrary to \textbf{\texttt{Random ID Label}} where the \textit{number} of entities is preserved, the number of entities in each annotation changes compared to the original ground-truth.

\noindent \textbf{\texttt{Random OOD Label}:} We replace entity labels in the ground-truth demonstrations with a random out-of-distribution~(OOD) English word.\footnote{\label{myfootnote}OOD words are randomly sampled from the English vocabulary in the \texttt{NLTK} library~\citep{bird2009natural}.}\addtocounter{footnote}{-1}\addtocounter{Hfootnote}{-1}

\noindent \textbf{\texttt{Random OOD Label from Text}:} We replace ground-truth entity labels with words randomly selected from the sample's text that are not included in the ground-truth annotation~(\ie, not a target entity).

\noindent \textbf{\texttt{Corrupted OOD Text}:} We replace the entity mentions in the text with random OOD English words.\footnotemark

\noindent \textbf{\texttt{Corrupted OOD Text and Label}:} Similar to \textbf{\texttt{Corrupted OOD Text}}, but we replace ground-truth labels as well, such that the entities in the text and label match.

\noindent \textbf{\texttt{Corrupted and Shuffled OOD Text}, \texttt{Corrupted and Shuffled OOD Text and Label}:} Same as their non-shuffled counterpart, but with randomly shuffling the words of the sentence.

\begin{figure*}[t]
    \centering
    \includegraphics[width=0.8\linewidth]{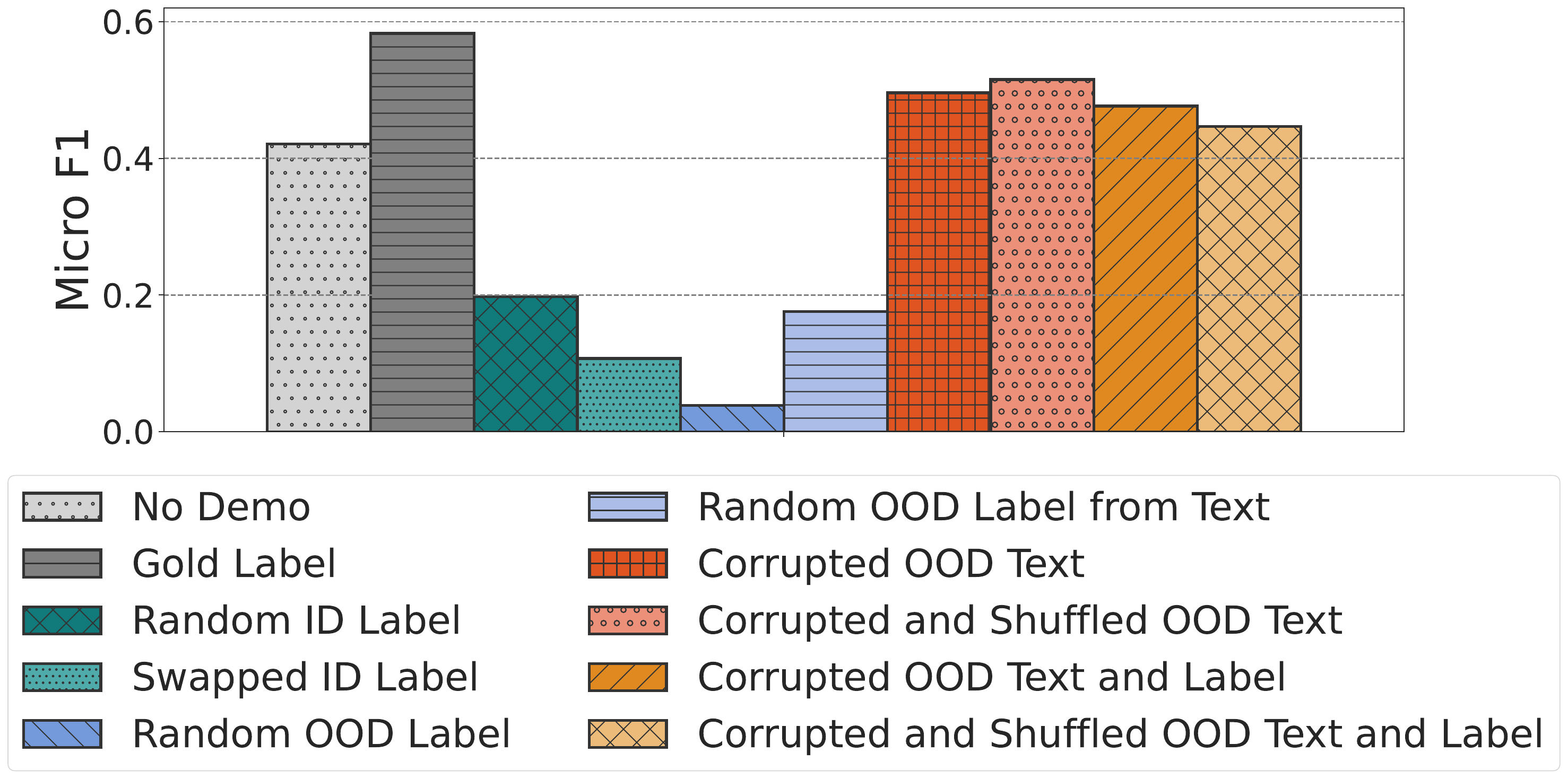}
    \caption{\textbf{10-shot ICL performance using various demonstration corruption schemes}, with Mistral and \knn{} demonstration retrieval. We compare to zero-shot and 10-shot with gold demonstrations, averaging over all datasets.}
    \label{fig:what_matters_in_icl_ner}
\end{figure*}


\paragraph{Results} 

When looking at Mistral's performance averaged over all datasets, we find as expected that demonstrations with gold annotations consistently improve the performance over no demonstration (Figure~\ref{fig:what_matters_in_icl_ner}).\footnote{For detailed results per dataset, and similar performance of \gptthree{}, see Appendix~\ref{app:analysis} (Figure~\ref{fig:perturb-all-datasets-gpt}).}

However, corrupting demonstrations lowers the performance, particularly for \textbf{\texttt{Random ID Label}}, \textbf{\texttt{Swapped ID Label}} and \textbf{\texttt{Random OOD Label}}, which are notably worse than zero-shot prediction. This observation differs from the findings of~\citet{min-etal-2022-rethinking} for in-context text classification and multiple choice QA, as well as~\citet{wang-etal-2023-towards}'s observations for QA with chain-of-thought reasoning, likely due to the open-endedness of the NED predictions (\ie, predicting multiple labels from a broad label space). In these three corruption schemes, the contextual and semantic correspondence between the input sentence and gold entities is broken.
As a result, the model learns spurious text-label correspondences from these demonstrations, leading to worse performance than in the zero-shot setting.
\textbf{\texttt{Random OOD Label from Text}} outperforms \textbf{\texttt{Random OOD Label}}, likely because it maintains a semantic correspondence between inputs and labels, but still underperforms zero-shot prediction due to misleading contextual associations.

Interestingly, both shuffled and unshuffled text corruption schemes (\textbf{\texttt{Corrupted (and Shuffled) OOD Text}}) exhibit no substantial performance drops, maintaining an edge over zero-shot prompting (despite the input prompt being the same for all corruptions in Table~\ref{tab:incorrect_demo_example}). 
We hypothesize that the model relies less on word order in the demonstrations to adapt to NED. Similar to how previous work showed that models no longer represent local word order in long contexts~\citep{sun-etal-2021-long}, we infer that the model does not need to represent explicit word order in exemplars to use them for transfer for a non-shuffled test sample.
Moreover, despite the label corruption in \textbf{\texttt{Corrupted (and Shuffled) OOD Text and Labels}}, the performance only slightly decreases compared to schemes with intact labels, still outperforming the zero-shot setting. This finding suggests that the model can induce entity presence from the global context, as ICL with these demonstrations still outperforms zero-shot predictions by up to $10\%$.

Based off these findings,  we conclude that for effective in-context task transfer in NED, the demonstrations must retain a degree of semantic correspondence between the input text and the extracted entities, but that the model's ability to adapt in-context is robust to noise in the demonstrations.

\begin{figure*}[t]
    \centering
    \includegraphics[width=1\linewidth]{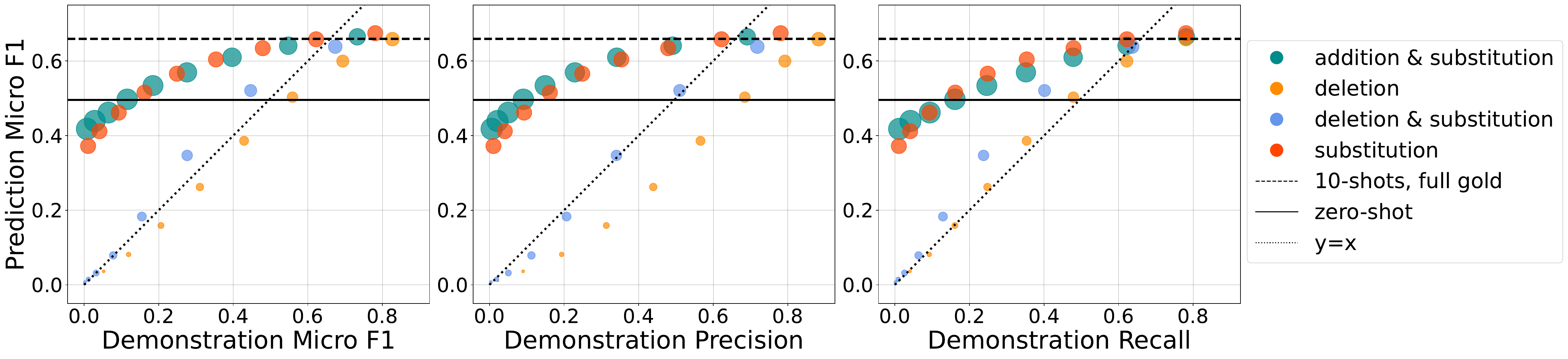}
    \caption{\textbf{10-shot ICL performance with perturbed demonstrations} with different perturbation schemes and using Mistral and \knn{} demonstration retrieval. We report the prediction F1 as a function of the precision, recall, and F1 of the perturbed demonstration label sets (relative to the gold demonstrations) averaged over all datasets. The size of the points shows the average number of entities in the label sets of the perturbed demonstrations.}
    \label{fig:perturbation}
    \vspace{-0.8em}
\end{figure*}

\subsection{Partially Correct ICL demonstrations}\label{sec:partial_correctness}
To further investigate the findings above, we perform a second study where we perturb demonstrations by modifying the context-label correspondence in a controlled manner. Specifically, we vary the correctness of the gold labels by applying different heuristic perturbations to the gold entity labels according to a perturbation factor $p \in \{0.1,0.2,...0.9\}$:

\noindent \textbf{\texttt{Deletion}}: each entity in the ground-truth annotation is deleted with probability $p$.

\noindent \textbf{\texttt{Substitution}}: each entity in the ground-truth annotation is substituted with a random in-distribution entity chosen from the dataset's label space with probability $p$.

\noindent \textbf{\texttt{Addition and Substitution}}: for each entity in the ground-truth, an entity chosen randomly from the dataset's label space is added to the annotation with probability $p$; additionally, each ground-truth entity is substituted with a random entity from the same label space with probability $p$.

\noindent \textbf{\texttt{Deletion and Substitution}}: each entity in the ground-truth is removed with probability $p$. The remaining entities are substituted with a random entity from the dataset with probability $p$.

\begin{figure*}[t]
    \centering
    \includegraphics[width=1.0\textwidth]{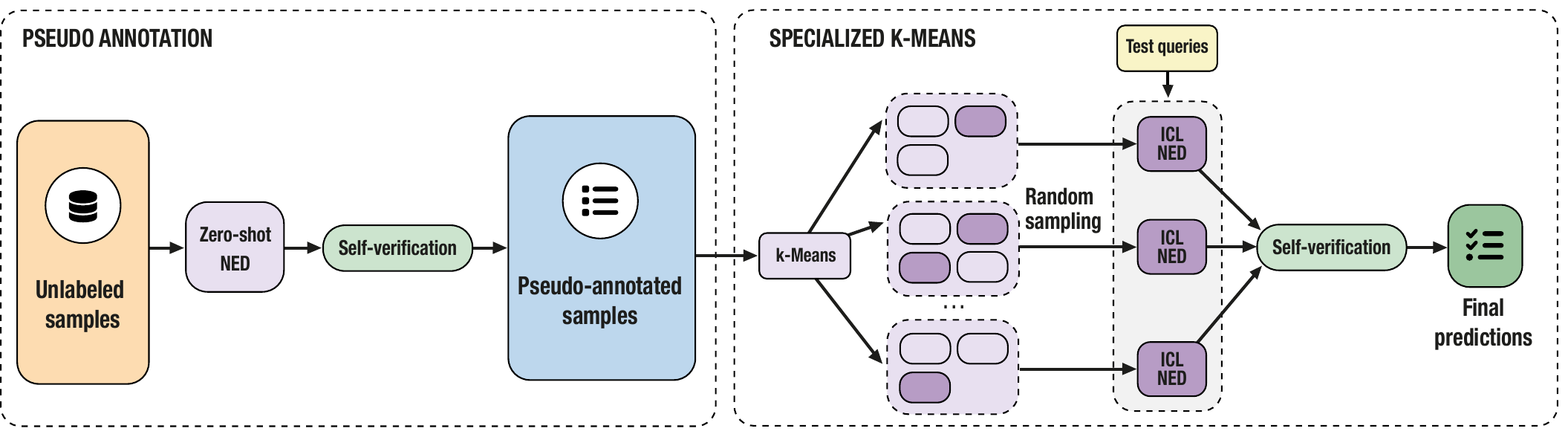}
    \caption{\textbf{\picl{} pipeline}. Unlabeled samples are pseudo-annotated through a zero-shot prediction and self-verification pass. Subsequently, they are clustered, and cluster-specific sets of ICL demonstrations are chosen at random from each group. Each set is independently used to find entity mentions in the query, and the final set of entity mentions is obtained by aggregating these independent sets and asking the model to verify the type of each predicted entity.}
    \label{fig:pseudo_ICL_methodology}
\end{figure*}

\noindent Following these perturbations, we report the precision, recall, and F1 score of the perturbed demonstrations~(evaluated based off the initial gold demonstration labels) against the F1 score of downstream predictions for test samples that contain at least one entity in their gold annotations.

\paragraph{Results} Demonstrations subject to different perturbations may have similar demonstration F1 scores, but result in considerably different prediction F1 scores~(Figure~\ref{fig:perturbation}). Specifically, we note that for a fixed demonstration F1 score, the perturbed demonstrations that retain a higher number of entities in the demonstration achieve much greater performance (\ie, \textbf{\texttt{Substitution}} and \textbf{\texttt{Addition and Substitution}}). Even with heavily perturbed demonstration labels, the prediction F1 stays above zero-shot performance and even remains close to the performance of 10-shot ICL (with gold labels) so long as some of the gold entities remain in the demonstration labels. Based off our findings, we hypothesize that \textbf{demonstrations with noisy but partially correct labels}~(such as those predicted by a zero-shot model) could benefit ICL for named entity detection.

Further results about the number of entities in the demonstration and perturbation factor, along with a comparison of the precision and recall of demonstrations against predictions, can be found in Appendix Figures~\ref{fig:app-perturbation-factor} and \ref{fig:app-perturbation-prc-rec}. These analyses show that for a given perturbation factor, changes that preserve or increase the total number of entities in the demonstration (such as \textbf{\texttt{Substitution}} or \textbf{\texttt{Addition and Substitution}}) cause a less pronounced performance drop.
\section{In-context NED with pseudo-annotated demonstrations}\label{sec:In-context NED with Pseudo-annotated Demonstrations}

In this section, we propose \picl, a framework for pseudo-annotating unlabeled samples that can be leveraged for in-context learning. This framework consists of two stages (Figure~\ref{fig:pseudo_ICL_methodology}). In the first stage, we start with a set of unlabeled samples and prompt the model in a zero-shot pass to extract entities in each sample.
Then, we improve the quality of these pseudo-annotations by prompting the model to verify each predicted entity (\ie, self-verification; ~\citealp{weng-etal-2023-large}), and filter entities that are not of the correct entity type. We use \kmeans{} clustering to group the pseudo-annotated samples into $K$ clusters based off the embedding of their text and pseudo-annotations.\footnote{We embed the text and entities of samples using the \texttt{S-PubMedBert-MS-MARCO} sentence transformer~(\url{https://huggingface.co/pritamdeka/S-PubMedBert-MS-MARCO}).} Each cluster is used as an individual pool of demonstrations for the downstream NED task. In the second stage, we prompt the model $K$ times, each time choosing the demonstrations from one cluster of pseudo-annotated samples (a sampling method we refer to as \spkmeans, \ie, Specialized \kmeans).
Then, for each entity in the $K$ lists of predictions, we perform a self-verification step to verify if the entity has the correct type or not, and retain the extracted entities that have the correct entity type. In all of our experiments, we pseudo-annotate $1000$ samples from the training set of the datasets with greedy decoding.
\begin{figure*}[t]
    \centering
    \includegraphics[width=1.0\textwidth]{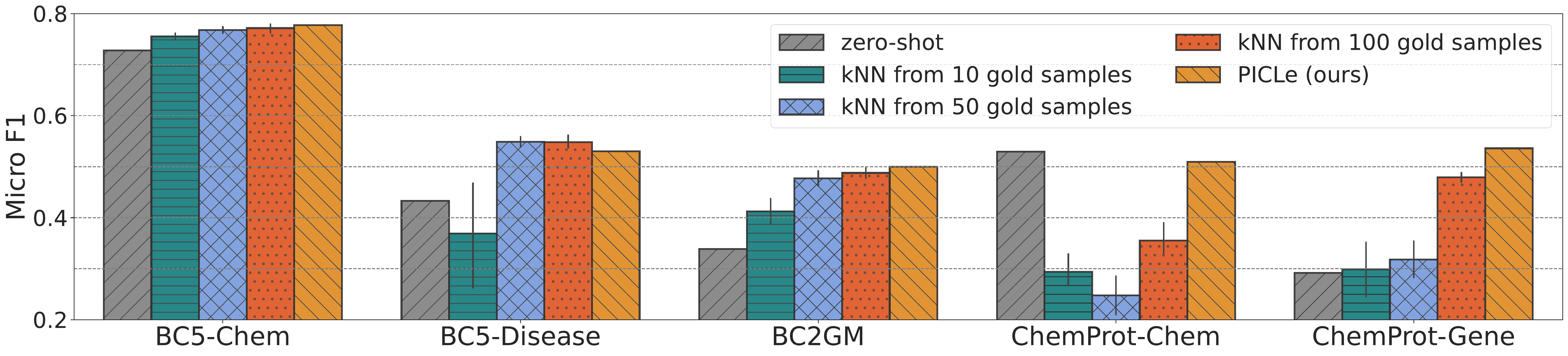}
    \caption{\textbf{Performance of \picl{}, zero-shot, and 10-shot ICL with gold demonstrations} selected from $10, 50, 100$ gold examples using Mistral. The error bars show the variance across $5$ seeds for sampling subsets of gold examples. All methods are followed by self-verification.} 
    \label{fig:pseudo-ICL-res}
\end{figure*}

\paragraph{\picl{} performance}

We evaluate \picl{} on the same biomedical NED datasets used for our analysis in Section~\ref{sec:What_Matters_in_In-context_NER} and compare \picl's performance with standard ICL using gold demonstrations sampled from different demonstration pool sizes, representing various degrees of annotation scarcity.
For baselines that use gold annotations as in-context examples, we initially sample demonstration pools of size $N$ from the full training set of each dataset, which range in size from $4.5$K to $12.5$K examples (Table \ref{tab:dataset_info}). In scarce annotation settings, we then sample demonstrations from these pools for gold in-context learning using \knn{} (following ablation study in Figure \ref{fig:ablation_gold} in Appendix). We experiment with $N \in [10,50,100]$, reporting results for a diverse set of annotation budgets. We repeat all experiments with $5$ seeds and report the average performance and standard deviation for these runs. 

Our results in Figure~\ref{fig:pseudo-ICL-res} show that across most datasets (with the exception of \textit{\chemprotchem}), \picl{} significantly outperforms the zero-shot baseline by an average of $10.7\%$ ($57.1\%$ compared to $46.4\%$). Furthermore, \picl{} also matches or outperforms in-context learning with gold demonstrations in resource-scarce settings, even beating an in-context learning baseline that has access to $100$ human-annotated demonstrations ($57.1\%$ vs. $52.8\%$). We note that the dataset with the highest performance, BC5chem ($77.7\%$ average F$1$ score for \picl), contains entity annotations whose surface forms generally contain fewer tokens (see Table \ref{tab:dataset_info}). On the contrary, the dataset with the lowest performance, BC2GM ($50\%$ F$1$ for \picl), has entity annotations that contain longer surface forms, making it more difficult to match the exact span in a generative manner.

We also compare \picl{} with a supervised baseline, fine-tuning a domain-specific language encoder, \texttt{BiomedNLP-BiomedBERT-large}, on various numbers of gold annotations (see Table \ref{tab:finetuning} in Appendix). While the performance of fine-tuning on $10$ gold samples is low and shows high variance between datasets, the performance with $50$ gold samples already outperforms all LLM baselines. However, we note that the sequence labeling formulation of the task for the supervised baseline differs from the generative formulation for LLMs, providing the supervised baseline with a simpler format for predicting entity spans, more adapted to our strict exact match evaluation.

\begin{table}[t]
    \centering
    \resizebox{\linewidth}{!}{
    \begin{tabular}{@{}rrllcc@{}}
\toprule
& \multicolumn{2}{c}{\textbf{Pseudo-annotation}}  & \multicolumn{2}{c}{\textbf{Inference}} &  \multirow{2}{*}{\textbf{F1}} \\ \cmidrule(lr){2-3} \cmidrule(lr){4-5}
   & \textbf{Runs} & \multirowcell{2}[0pt][c]{\textbf{Post-}\\ \textbf{processing}} &    \multirowcell{2}[0pt][c]{\textbf{Demo}\\ \textbf{retrieval}}  & \textbf{\texttt{SV}} & \\
& & & & & \\
\midrule
1 & 10 & merging &       &    \checkmark  &     55.7 \\
2 & 10 & \texttt{SV} &             \spkmeans   &    \checkmark           &           55.1 \\
3 & 1 & none &                &    \checkmark                &         51.8 \\ \midrule
        4 &  1 &  &            \knn      &    \checkmark           &         42.7 \\
         5 & 1 & \texttt{SV} &         random      &    \checkmark          &          47.9 \\
    6 &      1 &  &    \kmeans      &    \checkmark          &        55.1 \\ \midrule
 7 & 1 & \multirowcell{2}[0pt][l]{\texttt{SV}} &   \multirowcell{2}[0pt][l]{\spkmeans}      &    $\times$    &  49.2 \\
8 &     1 &  &     &    \checkmark  &      \textbf{57.1 }\\
\midrule
9 & \multirowcell{3}[0pt][l]{NA} &  \multirowcell{3}[0pt][l]{NA} & Zero-shot & $\times$ & 44.6 \\
10 &                             &                               & Zero-shot & \checkmark & 46.4 \\
11 &                             &                               & 10 Zero-shot & \checkmark & 50.6\\

\bottomrule
\end{tabular}}
    \caption{\textbf{Ablation of each component of \picl{}}, averaged over all $5$ datasets, using Mistral. \texttt{SV} refers to the use of self-verification and $10$ Zero-shot refers to $10$ zero-shot passes with a non-zero temperature.}
    \label{tab:ablation}
\end{table}

\begin{table}[t]
    \centering
    \begin{tabular}{@{}llc@{}}
        \toprule
        \textbf{Demonstration retrieval} & \textbf{Inference}   & \textbf{F1} \\
        \midrule
        \llama{} + \picl{}      & \multirow{4}{*}{\llama{}}  & 51.9 \\
         Zero-shot &   & 48.3 \\
        10 gold samples        &  & 45.5 \\
        100 gold samples  + \knn{}    &  &  53.6 \\
        \midrule
        Mistral + \picl{}  & \multirow{4}{*}{Mistral} & 57.1 \\
        Zero-shot & &  46.4 \\
        10 gold samples      &  & 42.6 \\
        100 gold samples    + \knn{}    &  & 52.8 \\
        Full train set      + \knn{} (oracle)    &  & 63.2 \\
        \midrule
        \gptthree{} + \picl{}  & Mistral  & 56.5 \\
        \bottomrule
    \end{tabular}
    \caption{\textbf{Performance of \picl{} using different LLMs for pseudo-annotation and prediction}, compared with zero-shot and $10$-shot ICL with gold annotations, averaged over all $5$ datasets.}
    \label{tab:other_models}
\end{table}

\paragraph{Ablation study}

In Table~\ref{tab:ablation}, we ablate each step of the \picl{} pipeline to evaluate the importance of each component (see Appendix Table \ref{tab:appendix-ablation} for a detailed version including precision and recall). 
For pseudo-annotation, similarly to \citet{wan-etal-2023-better}, we compare our method with running zero-shot prediction $10$ times with non-zero temperature~($T=0.7$) and filtering the $10$ sets of extracted entities using self-verification or merging (\ie, prompting the LLM to aggregate the entity lists). Both lead to a slightly lower F$1$ score (rows $\#1$ and $\#2$) while being much more computationally expensive than a single round of zero-shot prediction. We also find that self-verification helps with improving pseudo-annotations (row $\#3$ vs. row  $\#8$). Additional results demonstrating the effectiveness of self-verification in the pseudo-annotation and inference steps are presented in Tables~\ref{table:zero-shot-ablation-train} and~\ref{table:zero-shot-ablation-test}.

We also compare the effect of different demonstration retrieval methods: random sampling, \knn{}, vanilla \kmeans{}, and specialized \kmeans{} (\spkmeans{}).
\knn{} ($k=10$), known for being sensitive to noisy demonstrations \citep{zhang2022automatic}, scores the lowest (row $\#4$). 
For random retrieval, we sample demonstrations using $5$ different seeds; the predicted entity lists are merged and post-processed using self-verification (row $\#5$). Similarly, for \kmeans{} (row $\#6$), we randomly sample one demonstration per cluster, increasing the intra-run diversity. Conversely, in \spkmeans, demonstrations in each round are all sampled from the same cluster, maximizing inter-run diversity.
We sample demonstrations using $5$ different seeds, leading to $5$ inference runs. The predicted entity lists are merged and self-verified. The diversity of demonstrations for \kmeans{} leads to a higher recall than random ($48.6$ vs. $40\%$), but not as high as \spkmeans{} ($53.5\%$), which benefits from having separate clusters that lead to more varied predictions. Self-verification improves performance during inference (rows $\#7$ vs. $\#8$), especially precision ($+20\%$).

Finally, to validate the importance of pseudo-annotations on the downstream performance, we run zero-shot inference $10$ times with temperature $0.8$, and pool all predictions before applying self-verification (rows $\#9$-$\#11$ of Table~\ref{tab:ablation}). This baseline improves compared to vanilla zero-shot followed by self-verification, but underperforms \picl{}, showing that pseudo-annotations not only provide \textit{seeds} for diversity but also promote task transfer.
 
\paragraph{Model Generalization}

We measure the performance of \picl{} when using different models to generate pseudo-annotations: \llama{} and \gptthree{}. All experiments with gold samples are performed by randomly sampling demonstrations with $5$ seeds and averaging the results. On average over all datasets, \llama{} exhibits the same behavior as Mistral, outperforming zero-shot and 10-shot with scarce gold demonstrations (Table \ref{tab:other_models}, and Table \ref{tab:appendix-other_models} in Appendix). We note that for Mistral, \picl{} outperforms gold demonstrations in all resource-scarce settings (\ie, $N=100$). However, the performance is lower than gold demonstrations when we can sample demonstrations from the full training set (\ie, Full train set + \knn{}). We consider this performance to be an \textit{oracle} ICL setting given the large number of available training examples, which would not be realistic for low-resource settings, and might question the necessity of using ICL compared to fine-tuning a NED model.

Given the cost to use \gptthree{}, we only use it for the pseudo-labeling step of \picl{}. Since the pseudo-annotated samples are generated from the training set of each dataset, we measure the F$1$ score of samples pseudo-annotated by each model: Mistral's score is $45.8\%$, \llama{} reaches $48\%$, while \gptthree{} achieves $62.7\%$. 
Consequently, we experiment with using the pseudo-annotations from \gptthree{} as a demonstration pool for performing inference with Mistral, whose inference cost is lower. Using a higher-quality demonstration pool results in higher precision ($65.2\%$ vs. $61.8\%$) but comparable F$1$ score on average. This result is in line with our findings in Section \ref{sec:partial_correctness}, where the prediction F$1$ score curve becomes more flat with high values of demonstration F$1$ score~(Figure \ref{fig:perturbation}). More specifically, our inference method \spkmeans{} succeeds in leveraging the noisy demonstrations generated by Mistral, and slight improvements in the demonstration F$1$ score from using \gptthree{} demonstrations do not improve results.

\section{Conclusion}

In this work, we study the demonstration attributes that enable in-context generalization for named entity detection. We find that the context-label semantic correspondence is crucial for effective in-context NED, and without this correspondence, in-context examples hurt performance, pushing it below zero-shot NED. However, our analysis demonstrates that partially correct demonstration label sets are just as effective as gold label sets, provided a sufficient number of correct label mappings are found in the demonstration. Based off these findings, we design an ICL framework, \picl, for named entity detection that leverages LLMs to produce pseudo-annotated examples that can be used for in-context transfer. Our results on five biomedical NED datasets demonstrate that \picl{} is more effective than zero-shot prediction and outperforms in-context learning with gold demonstrations when gold demonstrations are scarce.

\section{Limitations}

\paragraph{Single Task.} This work introduces a method to alleviate annotation effort for named entity detection (NED) while achieving comparable performance to few-shot NED with human-labeled annotations. While this pipeline could be generalized to other tasks besides NED, the experiments presented in this paper are limited to this particular task. However, we demonstrate its effectiveness over a broad set of entity types. Similarly, further work is needed to generalize our conclusions on the partial correctness of demonstrations to all structured output tasks.

\paragraph{Sensitive applications.} We apply our system to documents from the biomedical domain. The evaluation sets are drawn from abstracts from published articles. However, the tools we develop can be used to extract the same type of entities in sensitive documents. Our tools were not tested for these applications, and practitioners should be aware that performance on such different types of documents is not guaranteed to transfer.

\paragraph{Annotation bias.} Annotated data can contain various forms of annotation bias, which lead trained models to make biased predictions when labeling entities based on the knowledge and beliefs of the annotators. This bias is usually alleviated following common annotation practices such as computing inter-rater agreement and having detailed annotation guidelines discussed with the annotators. However \picl{} only uses models' pseudo-annotations, since we focus on domains for which expert annotation is challenging to obtain. Consequently, given the lack of interpretability and training data openness of the used LLMs, we cannot assess the reliability and fairness of the demonstrations.

\section*{Acknowledgments}

We thank Said Gürbüz for his work on the fine-tuning experiment (\texttt{BioMedBERT}) in low-resource domains, Anna Sotnikova and Beatriz Borges for providing comments on earlier drafts of this paper, and Angelika Romanou for the illustration in Figure~\ref{fig:pseudo_ICL_methodology}. This project was supported by the EPFL Center for Imaging through a Collaborative Imaging Grant and a Boehringer Ingelheim Fonds PhD stipend (S.M.). We also gratefully acknowledge the support of Swiss National Science Foundation (No. 215390), Innosuisse (PFFS-21-29), Sony Group Corporation, and the Allen Institute for AI.

\bibliography{anthology,custom}

\appendix
\clearpage

\begin{figure*}[ht!]
    \centering
   \includegraphics[width=1\linewidth]{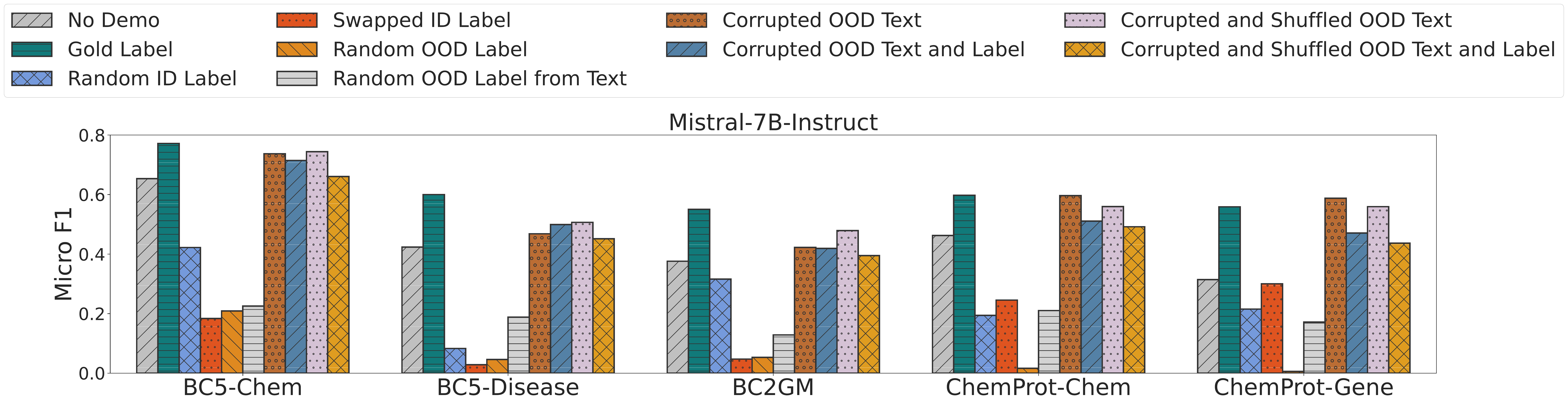}
   \includegraphics[width=1\linewidth, width=1\linewidth, trim={0 0 0 19em},clip]{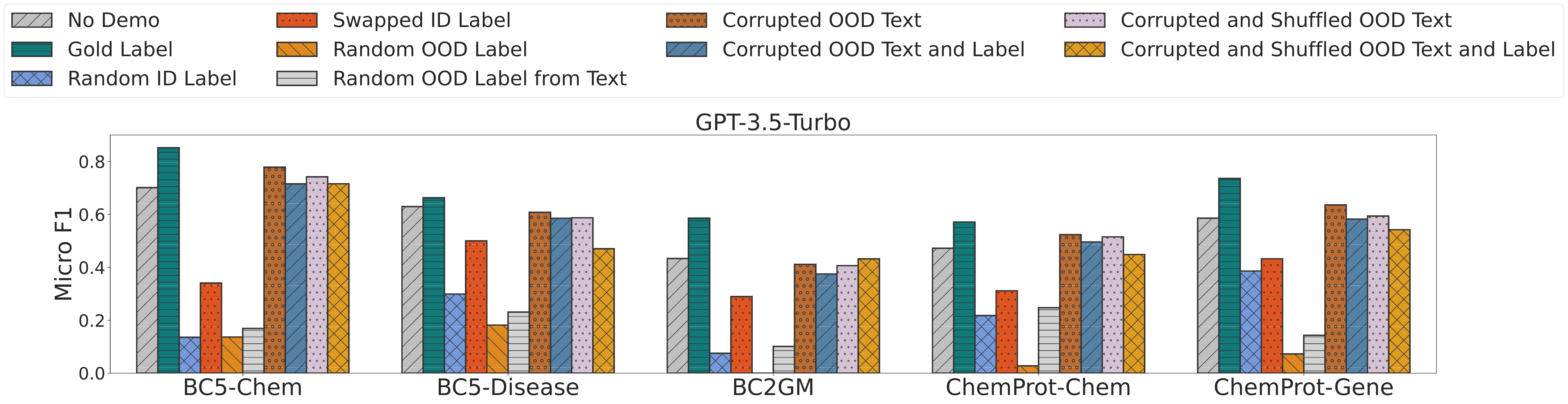}
    \vspace{-0.8em}
    \caption{\textbf{10-shot ICL performance using various demonstration corruption schemes}, compared with zero-shot and ICL with gold annotations, for each dataset. Experiments performed using Mistral~(top) and \gptthree{}~(bottom) and \knn{} demonstration retrieval. (Best viewed in color.)}
    \label{fig:perturb-all-datasets-gpt}
\end{figure*}

\begin{figure*}[ht!]
\centering
\begin{subfloat}
    \centering
    \includegraphics[width=1\linewidth]{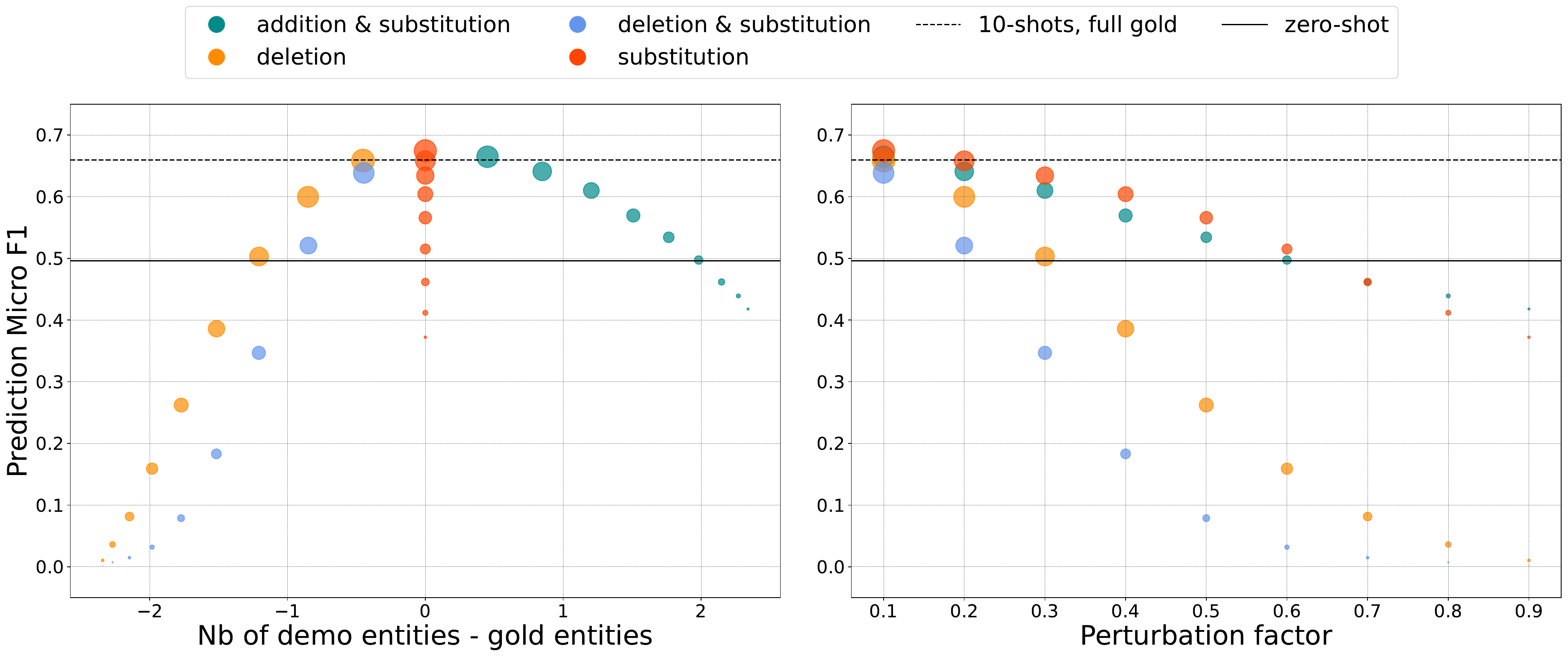}
    \caption{\textbf{10-shot ICL performance with partially correct demonstrations with different perturbation schemes} using Mistral and \knn{} demonstration retrieval. We observe the impact on prediction F1 score of the perturbation factor and the number of entities in the demonstrations for different perturbation types, averaged over all datasets. The size of the points shows the demonstrations' F1 score.}
    \label{fig:app-perturbation-factor}
\end{subfloat}%
\begin{subfloat}
    \centering
    \includegraphics[width=1\linewidth]{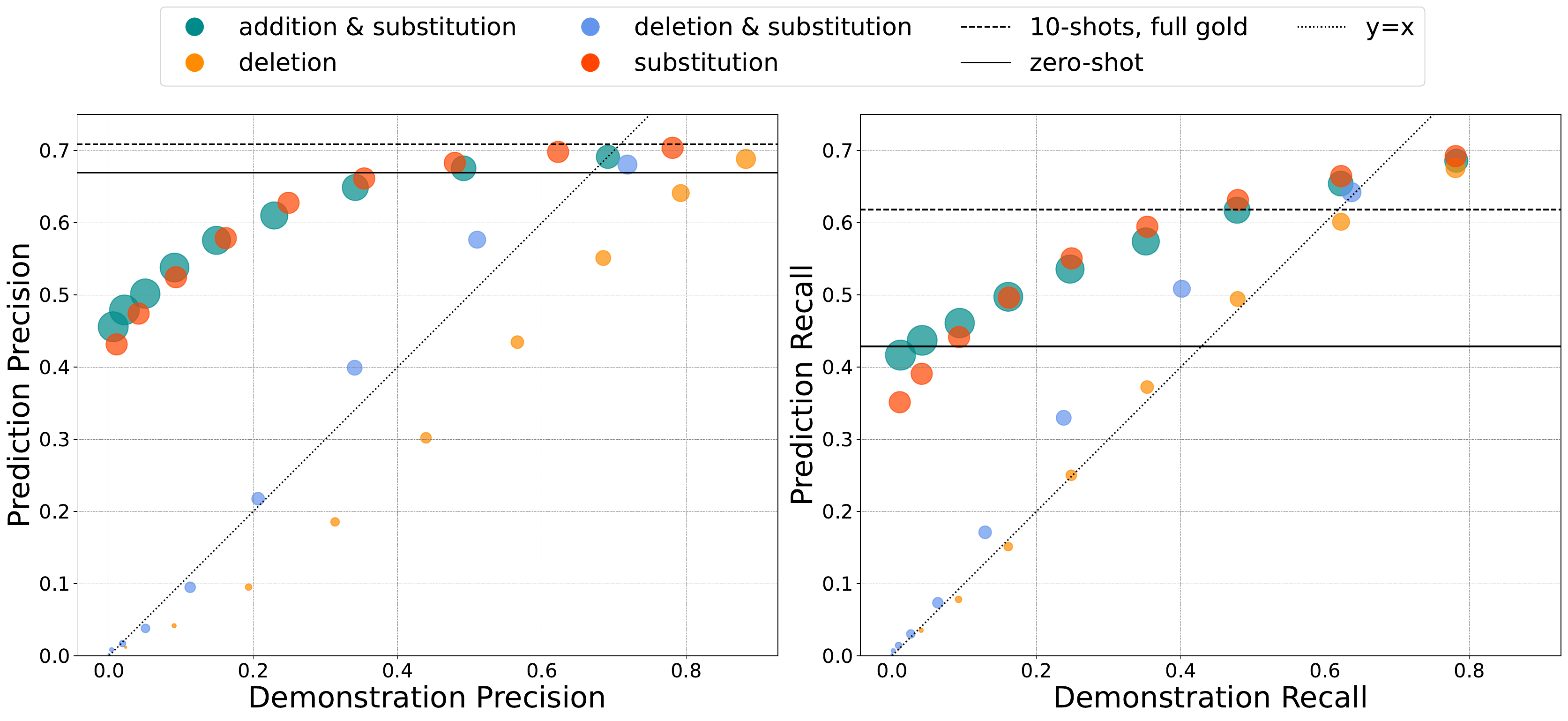}
    \caption{\textbf{10-shot ICL performance with partially correct demonstrations with different perturbation schemes} using Mistral and \knn{} demonstration retrieval. We observe the impact of the demonstration precision and recall on the downstream prediction precision and recall, respectively, averaged over all datasets. The size of the points shows the demonstrations' number of entities in the annotations.}
    \label{fig:app-perturbation-prc-rec}
\end{subfloat}
\end{figure*}

\section{Reproducibility statement}

\paragraph{Code.} We plan to share the code for 
\picl{} and all of our experiments. The decoding temperature is specified for each experiment in their corresponding section. For experiments with non-zero temperature, we use \texttt{top\_p}=$1$, and \texttt{max\_tokens}=$512$ in all experiments. All models were run on a single NVIDIA A100 GPU with 80 GB Memory, each inference run taking between 5 and 20 minutes depending on the dataset. 

\paragraph{Data.} The datasets we use are publicly available on the \texttt{Huggingface} platform.\footnote{\url{https://huggingface.co/datasets/bigbio/}}

\paragraph{Models.} As described in Section \ref{sec:exp-setup}, we use two open-source models for our studies whose checkpoints can be found in the \texttt{Huggingface} model library: Mistral-7b-instruct\footnote{\url{https://huggingface.co/mistralai/Mistral-7B-Instruct-v0.1}} and Llama-2-7b-Chat.\footnote{\url{https://huggingface.co/meta-llama/Llama-2-7b-chat-hf}} We also conduct experiments using a proprietary LLM from OpenAI, \texttt{gpt-3.5-turbo-0125},\footnote{\url{https://platform.openai.com/docs/models/gpt-3-5-turbo}} which unfortunately is subject to be updated (or removed from the API entirely) at any moment, limiting the long-term reproducibility of the results obtained with this tool. For supervised fine-tuning, we use the text encoder \texttt{BiomedNLP-BiomedBERT-large}.\footnote{\url{https://huggingface.co/microsoft/BiomedNLP-BiomedBERT-large-uncased-abstract}}

\paragraph{Random seeds}\label{app:seed}
We repeat all of our experiments that involve randomization 5 times with the following seeds: 12345, 24690, 37035, 49380, 61725.

\paragraph{Prompts}\label{app:prompts}

Examples of the prompts used for the self-verification pass and NED are shown in Figures~\ref{fig:prompt_self_ver} and \ref{fig:prompt_picle} respectively.

\section{Additional analysis for ICL demonstration}
\label{app:analysis}

\subsection{Corrupted random demonstrations}
Figure~\ref{fig:perturb-all-datasets-gpt} shows results per dataset for all corruption schemes with Mistral and \gptthree{}. We observe a similar trend for both models.

\subsection{Partially correct demonstrations}

Figure \ref{fig:app-perturbation-factor} shows the evolution of the downstream F1 score depending on the number of entities in the demonstrations and the perturbation factor. As expected, an increased perturbation factor leads to a lower demonstration F1 and a lower downstream F1 (right side of the figure). Similarly, adding or removing entities in the demonstration labels leads to a lower downstream F1.
However, with the same perturbation factor, perturbations that do not decrease the number of entities in the demonstration (\textbf{\texttt{Substitution}} and \textbf{\texttt{Addition and Substitution}}) lead to a much softer rate of performance loss. Similarly, to reach the same downstream performance as zero-shot (around $0.5$ on average), removing one entity is enough, while at least two entities need to be added.
This result supports the hypothesis that a way to increase downstream performance is to give preference to a higher recall and number of entities in the demonstration set.
Figure \ref{fig:app-perturbation-prc-rec} compares the precision and recall of demonstrations against the precision and recall of predictions.

\section{Additional results for \picl}
\label{app:results}

\paragraph{Ablation study}
We ablate each component of \picl{}'s pseudo-annotation and inference steps in Table~\ref{tab:appendix-ablation}. These results show the importance of the self-verification step in pseudo-annotation (row $\#3$ vs $\#8$), and inference (row $\#7$ vs $\#8$) for the downstream precision. Additionally, we observe that our proposed \spkmeans{} leads to a higher recall and F1 compared to other demonstration sampling methods (rows $\#4$-$6$ vs $\#8$).

\paragraph{Results with different models}
Table~\ref{tab:appendix-other_models} evaluates \picl{} with different base language models for pseudo-annotation and inference. We observe that for both \llama{} and Mistral, \picl{} obtains higher recall and F1 than zero-shot and almost all 10-shot with gold demonstration methods.

\paragraph{Impact of demonstration retrieval}

\begin{figure*}[ht]
    \centering
    \includegraphics[width=1\linewidth]{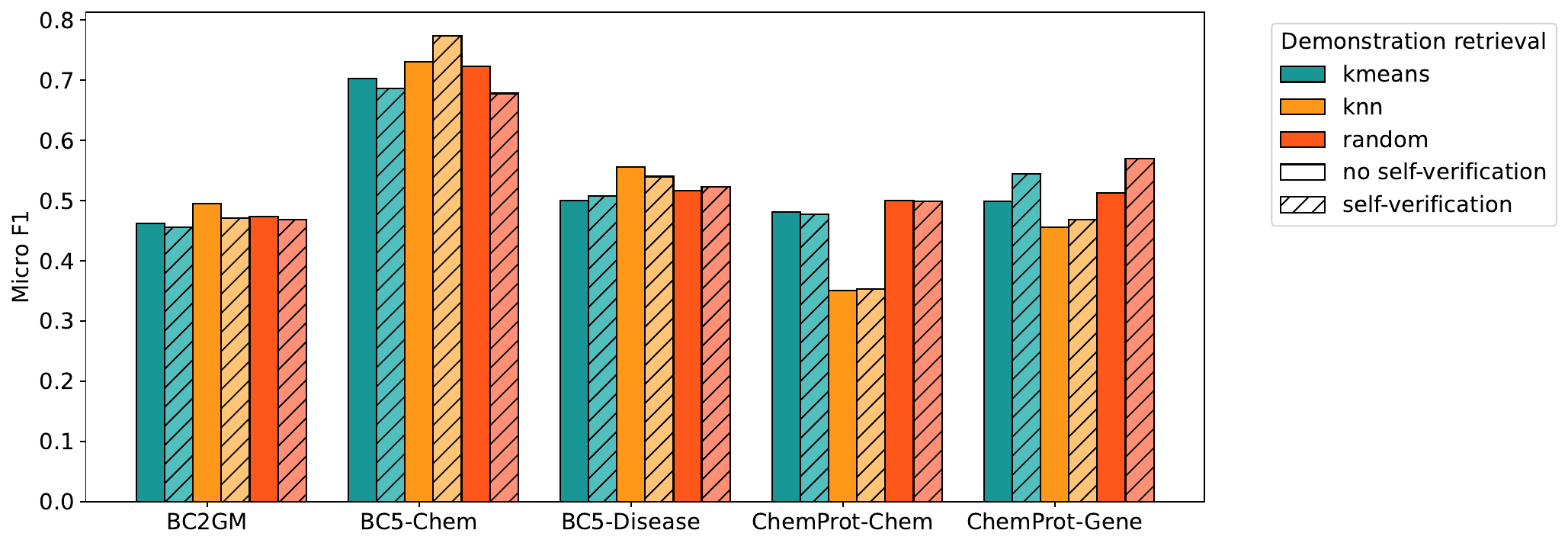}
    \caption{\textbf{10-shot NED with gold annotations} with Mistral using different demonstration retrieval methods: \kmeans, \knn, and random.}
    \label{fig:ablation_gold}
\end{figure*}

Here, we compare random, \kmeans, and \knn{} demonstration retrieval methods for gold demonstrations with and without the self-verification step. Figure~\ref{fig:ablation_gold} shows that demonstration retrieval performance between methods varies across datasets. However, \knn{} outperforms the other two methods on three out of five datasets. 

\begin{table}[t]
    \centering
    \resizebox{0.5\textwidth}{!}{
    \begin{tabular}{llccc}
    \toprule
    \textbf{Dataset} & \textbf{\texttt{SV}} & \textbf{Precision} & \textbf{Recall} & \textbf{Micro F1} \\
    \midrule
    \multirow{2}{*}{\bctwogm}         & $\times$   & 51.9 & 27.7 & 36.1 \\
                                      & \checkmark & 58.4 & 22.2 & 32.2 \\ 
    \midrule
    \multirow{2}{*}{\bcfivechem}      & $\times$   & 60.2 & 71.7 & 65.4 \\ 
                                      & \checkmark & 81.5 & 65.6 & 72.7 \\ 
    \midrule
    \multirow{2}{*}{\bcfivedisease}   & $\times$   & 57.1 & 33.3 & 42.1 \\ 
                                      & \checkmark & 69.9 & 31.3 & 43.2 \\ 
    \midrule
    \multirow{2}{*}{\chemprotchem}    & $\times$   & 34.6 & 54.7 & 42.4 \\ 
                                      & \checkmark & 53.1 & 49.9 & 51.5 \\ 
    \midrule
    \multirow{2}{*}{\chemprotgene}    & $\times$   & 69.1 & 20.7 & 31.9 \\ 
                                      & \checkmark & 77.6 & 18.3 & 29.6\\ 
    \midrule
    \multirow{2}{*}{\textbf{Average}} & $\times$   & 54.6 & 41.6 & 43.6 \\ 
                                      & \checkmark & 68.1 & 37.5 & 45.8 \\
\bottomrule
\end{tabular}}
\caption{\textbf{Evaluation of pseudo-annotated samples with and without self-verification.} The pseudo-annotations are obtained via zero-shot with Mistral with greedy decoding. \texttt{SV} refers to the use of self-verification.}
\label{table:zero-shot-ablation-train}
\end{table}

\begin{table}[t]
    \centering
    \resizebox{0.5\textwidth}{!}{
    \begin{tabular}{llccc}
    \toprule
    \textbf{Dataset} & \textbf{\texttt{SV}} & \textbf{Precision} & \textbf{Recall} & \textbf{Micro F1} \\
    \midrule
    \multirow{2}{*}{\bctwogm}          & $\times$   & 53.0 & 29.2 & 37.6 \\
                                       & \checkmark & 59.7 & 23.6 & 33.9  \\ 
    \midrule
    \multirow{2}{*}{\bcfivechem}       & $\times$   & 60.4 & 71.2 & 65.3 \\ 
                                       & \checkmark & 82.3 & 65.2 & 72.8 \\ 
    \midrule
    \multirow{2}{*}{\bcfivedisease}    & $\times$   & 53.9 & 34.9 & 42.4 \\ 
                                       & \checkmark & 67.3 & 31.9 & 43.3 \\ 
    \midrule
    \multirow{2}{*}{\chemprotchem}     & $\times$   & 39.4 & 56.1 & 46.3 \\ 
                                       & \checkmark & 56.7 & 49.7 & 53.0 \\ 
    \midrule
    \multirow{2}{*}{\chemprotgene}     & $\times$   & 68.8 & 20.4 & 31.4 \\ 
                                       & \checkmark & 77.4 & 18.0 & 29.2 \\ 
    \midrule
    \multirow{2}{*}{\textbf{Average}}  & $\times$   & 55.1 & 42.4 & 44.6  \\ 
                                       & \checkmark & 68.7 & 37.7 & 46.4 \\
\bottomrule
\end{tabular}}
\caption{\textbf{Evaluation of zero-shot inference with and without self-verification} using Mistral with greedy decoding. \texttt{SV} refers to the use of self-verification.}
\label{table:zero-shot-ablation-test}
\end{table}

\begin{table*}[t]
    \centering
    \begin{tabular}{lrllcrrr}
\toprule
& \multicolumn{2}{c}{\textbf{Pseudo-annotation}}  & \multicolumn{2}{c}{\textbf{Inference}} & \multirow{2}{*}{\textbf{Precision}} &  \multirow{2}{*}{\textbf{Recall}} &  \multirow{2}{*}{\textbf{F1}} \\ \cmidrule(lr){2-3} \cmidrule(lr){4-5}
   & \textbf{Runs} & \textbf{Post-processing} &    \textbf{Demo retrieval}  & \textbf{Self-verification} & & &  \\
\midrule
1 & 10 & LLM-merging  & \spkmeans &    \checkmark  & 56.7 & 55.1 & 55.7 \\
2 & 10 & self-verif   & \spkmeans &    \checkmark  & 56.3 & 54.2 & 55.1 \\
3 & 1  & none         & \spkmeans &    \checkmark  & 55.2 & 49.4 & 51.8 \\ 
\midrule
4 & 1  & self-verif   & \knn      &    \checkmark  & 72.5 & 32.8 & 42.7 \\
5 & 1  & self-verif   &  random   &    \checkmark  & 68.1 & 39.8 & 47.9 \\
6 & 1  & self-verif   & \kmeans   &    \checkmark  & 64.8 & 48.6 & 55.1 \\ 
\midrule
7 & 1  & self-verif   & \spkmeans &    $\times$    & 41.8 & 60.7 & 49.2 \\
8 & 1  & self-verif   & \spkmeans &    \checkmark  & 61.8 & 53.5 & \textbf{57.1 }\\
\bottomrule
\end{tabular}
    \caption{\textbf{Ablation of each component of \picl}, averaged over all datasets, using Mistral for pseudo-annotation and inference.}
    \label{tab:appendix-ablation}
\end{table*}

\begin{table*}[t]
    \centering
    \begin{tabular}{lllrrr}
        \toprule
        \textbf{Demonstration pool} & \textbf{Demo retrieval} & \textbf{Inference model}     & \textbf{Precision} & \textbf{Recall} & \textbf{F1} \\
        \midrule
        \picl{} (\llama)                & \spkmeans & \multirow{4}{*}{\llama{}}& 47.0 & 59.9 & 51.9 \\
                                        & zero-shot &                          & 59.5 & 40.8 & 48.3 \\
        10 gold samples                 &    -      &                          & 59.2 & 38.6 & 45.5 \\
        100 gold samples                & \knn{}                               &  & 60.0 & 48.7 & 53.6 \\
        \midrule
        \picl{} (Mistral)               & \spkmeans & \multirow{4}{*}{Mistral} & 61.8 & 53.5 & 57.1 \\
                                        & zero-shot &                          & 68.7 & 37.7 & 46.4 \\
        10 gold samples                 &   -       &                          & 65.7 & 34.9 & 42.6 \\
        100 gold samples                & \knn{}    &                          & 73.6 & 42.5 & 52.8 \\
        \midrule
        \picl{} (\gptthree{})           & \spkmeans & Mistral                  & 65.2 & 50.1 & 56.5 \\
        \bottomrule
    \end{tabular}
    \caption{ \textbf{Performance of PICLe using different LLMs for pseudo-annotation and prediction}, compared with zero-shot and 10-shot ICL with gold annotations, averaged over all 5 datasets.}
    \label{tab:appendix-other_models}
\end{table*}

\paragraph{Impact of self-verification}
We evaluate \picl's pseudo-annotations before and after self-verification and report the results per dataset in Table~\ref{table:zero-shot-ablation-train}.
We also report the performance of zero-shot NED before and after the self-verification pass in Table~\ref{table:zero-shot-ablation-test}.
It is observed that self-verification reduces the recall and significantly increases the precision, and on average leads to a higher F1 score, thereby improving the overall pseudo-annotation and prediction performance.

\paragraph{Fine-tuning results}

We report the performance of fine-tuned \texttt{BioMedBERT} in Table~\ref{tab:finetuning}. As expected with 10 gold samples fine-tuning results in low performance and high variance. However, the performance with 50 samples already outperforms all in-context learning baselines.

\paragraph{Efficiency analysis}
Compared to the $10$-shot baseline with gold demonstrations, \picl{} has a zero-shot pass followed by the self-verification step for pseudo-annotation, $10$ additional rounds in the \spkmeans{} component, and one final self-verification step. This is while \picl{} requires no human annotation which is crucial in low-resource domains.

\begin{table*}[t]
    \centering
    \resizebox{\textwidth}{!}{
    \begin{tabular}{lrrrrrr}
\toprule
\textbf{Train set size} &  \textbf{BC2GM} &  \textbf{\bcfivechem} &  \textbf{\bcfivedisease} &  \textbf{\chemprotchem }&  \textbf{\chemprotgene }&  \textbf{Average} \\
\midrule
\textbf{10}         &    9.8 &     59.0 &        15.1 &          64.3 &          50.1 &    39.7 \\
\textbf{50}         &   55.7 &     79.1 &        53.0 &          79.6 &          72.4 &    67.9 \\
\textbf{100 }       &   63.9 &     83.4 &        65.6 &          83.4 &          77.6 &    74.8 \\
\textbf{Full }      &   87.0 &     94.3 &        85.4 &          90.8 &          89.8 &    89.5 \\
\bottomrule
\end{tabular}}
   \caption{\textbf{Micro-F1 score of \texttt{BioMedBERT-large} fine-tuned on various numbers of gold annotations}. For 10, 50 and 100 gold annotations, random sets are sampled with 5 different seeds, and the fine-tuning performances are averaged.}
    \label{tab:finetuning}
\end{table*}


\newpage

\begin{figure*}[th!]
\begin{center}

\prompt{
\textbf{user}: Given the context and definition of chemical entity, answer the following question. Please reason about your answer and include YES or NO in your response. YES if the given phrase is a chemical entity, and NO if it is not. If you are not sure, you can say I don't know.\\
Context: p75NTR expression in rat urinary bladder sensory neurons and spinal cord with cyclophosphamide-induced cystitis.\\
Chemical definition: Chemical refers to any substance having a distinct molecular composition that is produced by or used in a chemical process. Chemicals can be elements or compounds, and they can exist in various forms—solid, liquid, or gas.\\
Based off this context and definition, does cyclophosphamide correspond to the name of a chemical entity?
}

\caption{Example of prompt used for \textbf{self-verification}. Dataset: BC5-Chem.}
\label{fig:prompt_self_ver}
\end{center}
\end{figure*}

\begin{figure*}[th!]
\begin{center}

\prompt{
\textbf{user}: Your task is to extract all of the chemicals mentioned in a given abstract published in PubMed.
Chemical refers to any substance having a distinct molecular composition that is produced by or used in a chemical process. Chemicals can be elements or compounds, and they can exist in various forms—solid, liquid, or gas.

Please extract all of the entities corresponding to chemicals from the following paragraph, the same way as they are marked in the examples.
Make sure to include all and only the chemicals mentioned in the text, but not the genes and proteins, and diseases. If there are no chemical entities in the text output 'None'. You will be penalized if you include an entity more or less than the number of times it appears in the text.\\

\#\# Here are some examples:\\

\textbf{user}: Input: No statistically significant changes in behavior or receptor binding were found in PD males with the exception of increased ( 3 ) H - MK - 801 binding in cortex.
Chemical entities:\\
\textbf{assistant}:\\
\texttt{H | yes | because it is a chemical\\
MK - 801 | yes | because it is a chemical}
\\
\\
\textbf{user}: Input: The normalized reflex amplitude was significantly higher during pain, but only at faster stretches in the painful muscle. Chemical entities:\\
\textbf{assistant}: 
\\
...
\\
\textbf{user}: Input: NRA0160 has negligible affinity for the human cloned dopamine D3 receptor ( Ki = 39 nM ) , rat serotonin ( 5 - HT ) 2A receptors ( Ki = 180 nM ) and rat alpha1 adrenoceptor ( Ki = 237 nM ).
Chemical entities:\\
\textbf{assistant}: \\
\texttt{NRA0160 | yes | because it is a chemical\\
dopamine | yes | because it is a chemical\\
serotonin | yes | because it is a chemical\\
5 - HT | yes | because it is a chemical\\}

\textbf{user}: Input: p75NTR expression in rat urinary bladder sensory neurons and spinal cord with cyclophosphamide-induced cystitis. Chemical entities:}

\end{center}
\caption{Example of prompt used for \textbf{NED}. Dataset: BC5-Chem.}
\label{fig:prompt_picle}
\end{figure*}

\end{document}